\documentclass[11pt]{article}

\usepackage[preprint]{acl}
\usepackage{times}
\usepackage{latexsym}

\usepackage[T1]{fontenc}

\usepackage[utf8]{inputenc}

\usepackage{microtype}

\usepackage{inconsolata}
\usepackage{shorttoc}
\usepackage{hyperref} 
\usepackage{amsmath} 
\usepackage{algorithm}
\usepackage{algpseudocode}

\usepackage{graphicx}
\usepackage{booktabs,adjustbox, multirow} 
\usepackage{siunitx} 
\usepackage{xcolor}
\usepackage[most]{tcolorbox}
\usepackage[inline]{enumitem}
\usepackage{lmodern}
\usepackage{lipsum}  

\hypersetup{
    colorlinks=true,
    linkcolor=blue,
    urlcolor=blue
}

\newtcblisting{promptlisting}{
    listing only,
    breakable,
    enhanced,
    colback=white,
    colframe=white,
    boxrule=0pt,
    left=0pt,
    right=0pt,
    top=0pt,
    bottom=0pt,
    listing options={
        basicstyle=\ttfamily\small,
        breaklines=true,
        columns=fullflexible,
        keepspaces=true,
        showstringspaces=false
    }
}

\title{PAR$^2$-RAG: Planned Active Retrieval and Reasoning for Multi-Hop Question Answering}

\author{Xingyu Li,~Rongguang Wang,~Yuying Wang,~Mengqing Guo,~\\ {\bf Chenyang Li,~Tao Sheng,~Sujith Ravi,~Dan Roth} \\
        \bf{Oracle AI} }

\begin{document}
\maketitle
\begin{abstract}
Large language models (LLMs) remain brittle on multi-hop question answering (MHQA), where answering requires combining evidence across documents through retrieval and reasoning. Iterative retrieval systems can fail by locking onto an early low-recall trajectory and amplifying downstream errors, while planning-only approaches may produce static query sets that cannot adapt when intermediate evidence changes. We propose \textbf{Planned Active Retrieval and Reasoning RAG (PAR$^2$-RAG)}, a two-stage framework that separates \emph{coverage} from \emph{commitment}. PAR$^2$-RAG first performs breadth-first anchoring to build a high-recall evidence frontier, then applies depth-first refinement with evidence sufficiency control in an iterative loop. Across four MHQA benchmarks, PAR$^2$-RAG consistently outperforms existing state-of-the-art baselines, compared with IRCoT, PAR$^2$-RAG achieves up to \textbf{23.5\%} higher accuracy, with retrieval gains of up to \textbf{10.5\%} in NDCG.
\end{abstract}

\section{Introduction}
Recent LLMs have achieved expert-level performance in domains such as mathematics and coding~\cite{openai2023gpt4,deepseek2025r1,team2024gemini15,openai2025o3}, yet they still struggle with multi-hop question answering (MHQA), where systems must gather evidence from multiple documents and compose those facts into a coherent answer~\cite{ho2020constructing,trivedi2022musique,schnitzler2024morehopqa,krishna2025fact}. Retrieval-Augmented Generation (RAG) and related approaches were designed to address this gap, but they still fail frequently in multi-hop settings when evidence coverage is incomplete or reasoning trajectories drift~\cite{lewis2020retrieval,gao2023retrieval,trivedi2023interleaving}.


A common failure mode in MHQA is \emph{premature commitment}, where retrieval approaches interleave reasoning with retrieval in a mostly greedy, depth-first manner~\cite{yao2023react,trivedi2023interleaving}. When early steps latch onto a distractor, later hops amplify that error and recovery becomes difficult. Meanwhile, planning-heavy methods can produce broad but static query sets that become misaligned when intermediate evidence changes~\cite{cheng2025dualrag,liu2025hoprag}.

This motivates a central question: \emph{can we improve multi-hop robustness by jointly controlling retrieval coverage and the timing of reasoning commitment?} 
We propose \textbf{Planned Active Retrieval and Reasoning RAG (PAR$^2$-RAG)}, an agentic and modular two-stage evidence search approach that separates \emph{retrieval coverage} from \emph{reasoning commitment}. Stage~1 performs \emph{coverage anchor} to expand the evidence pool. Stage~2 performs \emph{tterative chain refinement} within that anchored pool to construct a coherent retrieval evidence chain with intermediate sub-thoughts. The generation module then conditions on this structured evidence, which improves final answer quality by making synthesis more evidence-grounded. 

We evaluate PAR$^2$-RAG on four MHQA benchmarks using multiple answer and retrieval quality metrics. PAR$^2$-RAG consistently outperforms strong training-free baselines on both retrieval coverage and answer accuracy, with 23.5\% gain over IRCoT in accuracy and retrieval gains of10.3\% in recall and 10.5\% in NDCG. 

We make three primary contributions:
\begin{itemize}
    \item We introduce PAR$^2$-RAG, an agentic modular two-stage framework for MHQA through planned active retrieval and evidence-aware reasoning.
    \item We provide empirical study under both non-reasoning and reasoning settings, including reasoning-intensive MHQA benchmarks, showing consistent gains over strong training-free baselines.
    \item We present mechanism-level diagnostics that connect answer gains to retrieval behavior, together with robustness analyses.
\end{itemize}

\section{Related Work}

\subsection{Retrieval-Augmented Generation}
Retrieval-Augmented Generation (RAG) improves factual grounding by coupling generation with external evidence retrieval~\cite{lewis2020retrieval,guu2020retrieval}. Follow-up work improves retrieval quality through dense encoders, re-ranking, and retrieval-aware prompting~\cite{karpukhin2020dense,izacard2021unsupervised,shi2023replug,izacard2022few,asai2024self}. However, strong first-hop retrieval does not by itself solve multi-hop reasoning: systems can still fail when evidence is incomplete or poorly composed across steps~\cite{gao2023retrieval,tang2024multihop}.

\subsection{Training-free Multi-Hop QA}
Interleaving retrieval with reasoning is a dominant training-free strategy for multi-hop question answering (MHQA). ReAct alternates thought and action to iteratively collect evidence~\cite{yao2023react}, and IRCoT interleaves chain-of-thought with retrieval to improve compositional QA performance~\cite{trivedi2023interleaving}. These methods are flexible and often strong, but retrieval and commitment are coupled in a single loop. Recent analyses of multi-hop behavior and retrieval difficulty indicate that errors in early hops can propagate and degrade later reasoning, especially when bridge evidence is fragile or missing~\cite{biran2024hopping,zhu2025mitigating}.

Other RAG-based approaches improve MHQA by decomposing questions into sub-queries and searching broader evidence frontiers before deep reasoning. Decomposition and structured retrieval can improve coverage, especially for bridge-fact questions, but purely static plans can become mismatched when intermediate evidence shifts the information need. Recent modular and structured pipelines (e.g., explicit retrieval-in-context search or retrieval--inference coupling) make this trade-off between breadth and commitment control explicit~\cite{chen-etal-2025-search,li2024brief,cheng2025dualrag,liu2025hoprag}.

\section{Methods}
\begin{figure*}[t!]
\centering
\includegraphics[width=0.8\linewidth]{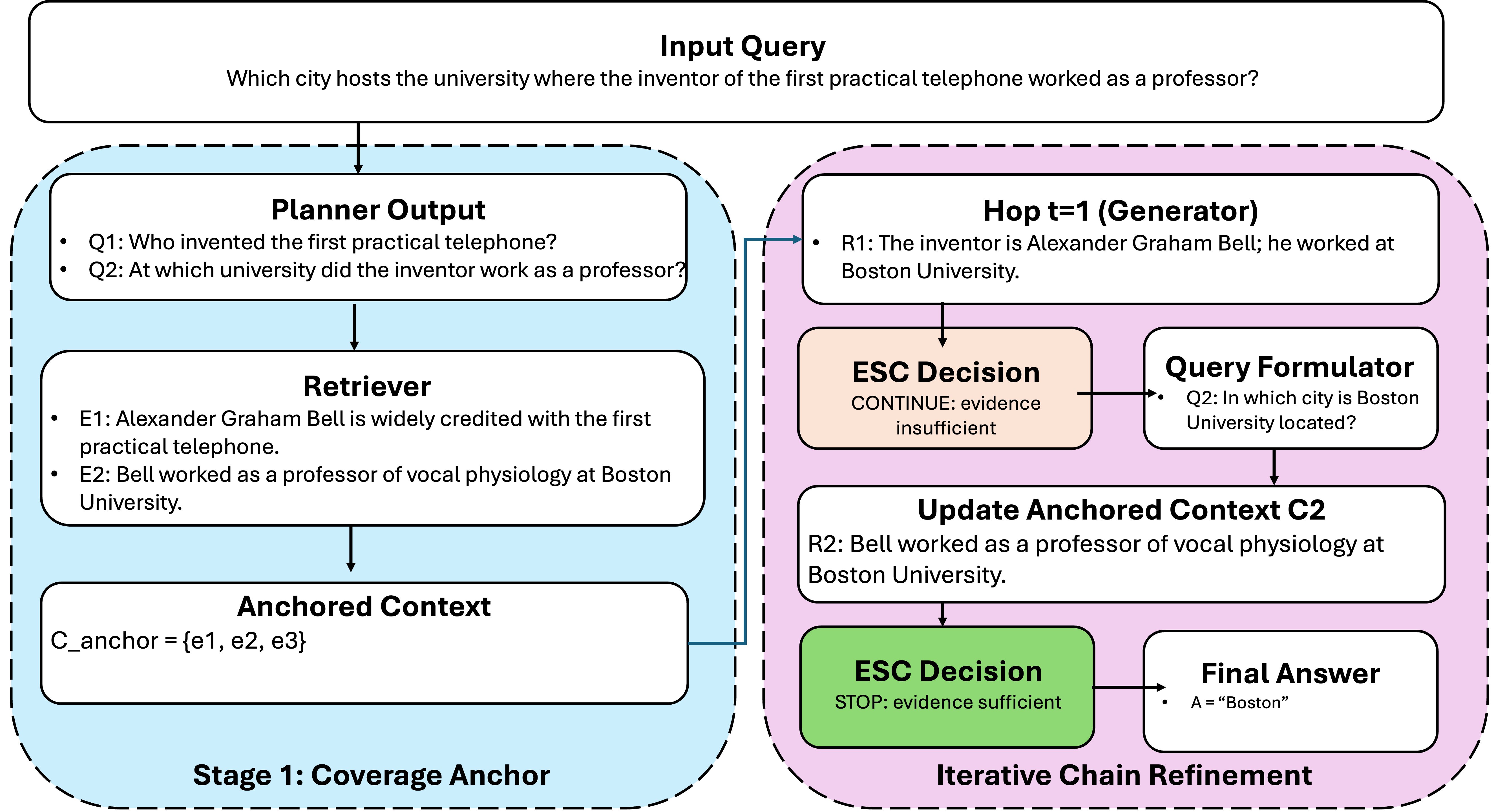}
\caption{Overall architecture of PAR$^2$-RAG. Stage~1 (Coverage Anchor) expands evidence breadth to build $C_{\text{anchor}}$, and Stage~2 (Iterative Chain) performs ESC-gated refinement to either continue targeted retrieval or stop with the final answer.}
\label{fig:system_architec}
\end{figure*}
Multi-hop QA systems often fail when they commit to a reasoning path before evidence coverage is sufficient. This premature commitment can lock retrieval into low-recall trajectories and amplify downstream errors. Our goal is to address this failure mode by explicitly controlling \emph{when} the system commits to reasoning and \emph{how} retrieval is expanded beforehand. We therefore propose \textbf{Planned Active Retrieval and Reasoning RAG (PAR$^2$-RAG)}, a two-stage framework that separates \emph{coverage acquisition} from \emph{reasoning commitment}. The key design principle is \emph{coverage first, commitment late}: first build a broad evidence frontier, then perform controlled chain refinement with explicit sufficiency checks.

\subsection{PAR$^2$-RAG}
To present PAR$^2$-RAG, we first formulate the multi-hop RAG. Let $q$ denote the input multi-hop question and $D=\{d_1,\dots,d_N\}$ the document corpus, where each $d_n$ is a chunked passage. A multi-hop RAG process can be viewed as iterative sub-question planning, retrieval, and response generation. Within this setting, PAR$^2$-RAG uses five agents: (1) a Planner $\mathcal{P}(q)$ that generates coverage-oriented sub-queries to expand the evidence frontier; (2) a Retriever $\mathcal{R}(q,D,k)$ that returns top-$k$ ranked passage evidence $E_i$ for a query $q_i$; (3) a Query Formulator $\mathcal{Q}$ that rewrites follow-up queries for commitment-stage refinement; (4) an Evidence Sufficiency Controller (ESC) $\mathcal{E}$ that decides whether to continue retrieval via $\mathcal{Q}$ or stop and finalize the answer; and (5) Writer $\mathcal{W}$ to generate the intermediate response or the final answer with prompt-instructed LLM calls.

Figure~\ref{fig:system_architec} provides an overview of PAR$^2$-RAG. The architecture has two coordinated stages. Stage~1 (Coverage Anchor) performs breadth-first evidence expansion: the planner $\mathcal{P}(q)$  decomposes the query into complementary sub-queries, the retriever $\mathcal{R}$ gathers candidate passages, and the system merges them into an anchored context $C_{\text{anchor}}$. Stage~2 (Iterative Chain) performs depth-first refinement over this anchored context. At each hop, the writer $\mathcal{W}$ produces a step response and the ESC $\mathcal{E}$ decides whether to continue retrieval with a reformulated query or finalize the answer. This control loop makes retrieval adaptive to missing evidence while preventing premature commitment to an incomplete reasoning path. 

\paragraph{Stage 1: Coverage Anchoring }
Stage~1 implements the \textbf{Coverage Anchor} module, whose objective is to construct a high-recall evidence frontier before the system commits to a narrow reasoning path. This stage uses two agents: a planner $\mathcal{P}(q)$ that produces diverse decomposed sub-queries, and a retriever--reranker $\mathcal{R}$ that executes these queries and returns ranked, deduplicated passages. Note that both agents are necessary: decomposed queries define retrieval intents, but they do not provide factual grounding by themselves. The retrieved chunks instantiate an anchor context $C_{\text{anchor}}$ grounded in corpus evidence. This grounding reduces premature commitment, improves bridge-fact coverage, and provides a stable initialization for Stage~2 refinement.

Concretely, given $q$, the planner generates a diverse set of $m$ sub-queries $\{q_i\}_{i=1}^m$ targeting complementary aspects of the information need. For each $q_i$, the system retrieves top-$k$ candidates, reranks, deduplicates, and merges the retained passages into $C_{\text{anchor}}$. The resulting anchored context serves as a global evidence frontier, increasing the likelihood that required supporting facts are available before deep reasoning begins.

\paragraph{Stage 2: Iterative Chain Refinement}

Stage~2 corresponds to the \textbf{Iterative Chain} module, which aims to convert broad anchored evidence into a precise, answer-supporting reasoning chain. In contrast to Stage~1, this stage prioritizes commitment quality: deciding whether current evidence is sufficient for the next reasoning step, or whether targeted retrieval still needs refinement. The \textbf{Iterative Chain} stage uses three complementary agent: a generator $\mathcal{G}$ to produce step-level responses from the current context, a query formulator $\mathcal{Q}$ to propose focused follow-up query formulation when evidence is incomplete, and an ESC $\mathcal{E}$ to decide whether to query reformulation or answer generation. Considering at hop $t$, the generator $\mathcal{G}$ a step response $r_t$ from the current context $C_t$ (when t=1, $C_1$ is equal to $C_{\text{anchor}}$), then $\mathcal{E}$is invoked to decide the next step as:
\begin{equation}
\mathcal{E}(q, r_t, C_t) \rightarrow (\texttt{action},\, q_{t+1}^{\star}),
\end{equation}
where $\texttt{action}\in\{\texttt{CONTINUE},\texttt{STOP}\}$ and a prompt-instructed LLM makes the decision for $\mathcal{E}$. If $\mathcal{E}$ outputs \texttt{CONTINUE}, the system issues a reformulated query $q_{t+1}^{\star}$ and retrieves additional evidence for $C_{t+1}$; if $\mathcal{E}$ outputs \texttt{STOP}, the loop terminates and returns the answer.

Algorithm~\ref{alg:parrag} summarizes the implementation of PAR$^2$-RAG. After stage 1, at each hop $t$, $\mathcal{G}$ produces $r_t=\mathcal{G}(q,C_t)$, then ESC returns $(\texttt{action}, q_{t+1}^{\star}, m_t)$. If $\texttt{action}=\texttt{CONTINUE}$, the generated $q_{t+1}^{\star}$ from $\mathcal{Q}$ invokes $\mathcal{R}$, and updates context to $C_{t+1}$; otherwise the loop terminates and returns the latest response as the final answer. This keeps retrieval adaptive while explicitly controlling continuation to reduce premature commitment and off-path reasoning.

\begin{algorithm}[t!]
\caption{PAR$^2$-RAG}
\label{alg:parrag}
\begin{algorithmic}[1]
\Require Query $q$, corpus $D$, hop budget $H$
\Ensure Final answer $a$
\State $\{q_i\}_{i=1}^m \gets \mathcal{P}(q)$
\State Initialize $C_{\text{anchor}}$
\For{each $q_i$}
    \State $E_i \gets \mathcal{R}(q_i,D,k)$
    \State $C_{\text{anchor}} \gets C_{\text{anchor}} \cup E_i$
\EndFor
\State $t \gets 1$, $C_1 \gets C_{\text{anchor}}$
\While{$t \le H$}
    \State $r_t \gets \mathcal{G}(q, C_t)$
    \State $(\texttt{action}, q_{t+1}^{\star}, m_t) \gets \mathcal{E}(q, r_t, C_t)$
    \If{$\texttt{action}=\texttt{STOP}$}
        \State \textbf{break}
    \EndIf
    \State $E_{t+1} \gets \mathcal{R}(q_{t+1}^{\star},D,k)$
    \State $C_{t+1} \gets C_t \cup E_{t+1}$
\EndWhile
\State $a \gets r_t$
\State \Return $a$
\end{algorithmic}
\end{algorithm}

\subsection{Discussion}
PAR$^2$-RAG is designed around a simple control principle: \emph{retrieve broadly before committing narrowly}. Different from RAG methods such as ReAct and IRCoT, PAR$^2$-RAG separates coverage expansion from commitment-time refinement, so early low-recall evidence is less likely to lock the system into an off-path trajectory. Comparing to planning-only decomposition, PAR$^2$-RAG remains adaptive during refinement through ESC-guided continuation and query reformulation.

This agentic decomposition also clarifies the role of each component. Coverage Anchor increases evidence breadth and bridge-fact recall; Iterative Chain improves local trajectory quality; ESC decides when additional retrieval is necessary versus when evidence is sufficient to stop. Compared with single-module variants, the combined design offers complementary benefits: stronger initial recall, controlled correction when evidence is incomplete, and more stable behavior across multi-agents.

\section{Experiments}\label{sec:experiments}

\subsection{Experimental Setups}
\paragraph{Benchmarks}
We evaluate PAR$^2$-RAG on four benchmarks spanning two categories: 2WikiMultiHopQA~\cite{ho2020constructing} and MuSiQue~\cite{trivedi2022musique}, which emphasize connected multi-step evidence composition; and MoreHopQA~\cite{schnitzler2024morehopqa} and FRAMES~\cite{krishna2025fact}, which further stress generative reasoning and end-to-end retrieval-grounded factuality. For each dataset, we sample 500 queries from the validation split. Dataset details are provided in Appendix~\ref{appendix:dataset}.

\paragraph{Baselines}
To provide evaluation of PAR$^2$-RAG, we compare against three-types of training-free methods: (1) \emph{Non-retrieval} baselines: Direct Inference and Chain-of-Thought (CoT)~\cite{wei2022chain}. (2) \emph{Iterative retrieval--reasoning} baselines: ReAct~\cite{yao2023react} and IRCoT~\cite{trivedi2023interleaving}, which we treat as strong training-free RAG baselines. (3) \emph{PAR$^2$-RAG module variants}: Coverage Anchor and Iterative Chain, which correspond to modular components of PAR$^2$-RAG rather than independent external methods.

Evaluating these two module variants separately helps clarify \emph{why} PAR$^2$-RAG works: (a) it isolates gains from planning-oriented coverage expansion versus iterative chain refinement, (b) it reveals whether errors come from weak global evidence coverage or weak local chain updates, and (c) it quantifies the complementarity between planning and reasoning-chain control when both modules are combined.

\paragraph{Evaluation Metrics}
We evaluate both answer quality and retrieval quality. For answer quality, we use binary correctness judged against ground truth by OpenAI GPT-5-mini~\cite{singh2025openaigpt5card}. For retrieval quality, we compare deduplicated retrieved chunk IDs against ground-truth documents using Recall@k, NDCG@k~\cite{jeunen2024normalised}, and All-Pass (1 only when all required documents are retrieved). We report query-level averages and per-required-length diagnostics. Full metric definitions are provided in Appendix~\ref{appendix:metrics}.

\paragraph{Implementation Details}
Our pipeline follows ingestion, retrieval, and generation stages. During ingestion, each benchmark corpus is split into chunks and indexed in OpenSearch with 1024-dimensional embeddings. At inference time, retrieval uses two stages: broad hybrid candidate retrieval (TopK=500) followed by reranking with \texttt{e5-mistral-7b-instruct}~\cite{wang2024improving} to produce a compact evidence set (TopK=5). We use GPT-4.1 for the Coverage Anchor and Iterative Chain agents, and GPT-o4-mini as the evidence controller. To test whether retrieval-policy gains persist under stronger generation-time reasoning, we evaluate both non-reasoning and reasoning generation settings using GPT-4.1 and GPT-o3, respectively. Detailed settings appear in Appendix~\ref{appendix:implementation}.

\subsection{Main Results}
\begin{table}[t!] 
\centering
\begin{adjustbox}{width=0.98\columnwidth,center}
\begin{tabular}{@{}l c c c c c@{}}
\toprule
\textbf{Method} & \textbf{MuSiQue} & \textbf{2Wiki} & \textbf{MoreHopQA} & \textbf{FRAMES} & \textbf{Average} \\
\midrule
Direct Inference & 0.204 & 0.418 & 0.144 & 0.294 & 0.265 \\
CoT & 0.338 & 0.724 & 0.594 & 0.631 & 0.572 \\
\midrule
ReAct & 0.432 & 0.724 & 0.740 & 0.745 & 0.660 \\
IRCoT & 0.498 & 0.820 & 0.754 & 0.726 & 0.700 \\
\midrule
Coverage Anchor & \underline{0.539} & 0.835 & 0.753 & 0.774 & 0.725 \\
Iterative Chain & 0.516 & \underline{0.868} & \underline{0.800} & \underline{0.788} & \underline{0.743} \\
PAR$^2$-RAG & \textbf{0.615} & \textbf{0.896} & \textbf{0.826} & \textbf{0.811} & \textbf{0.787} \\
\bottomrule
\end{tabular}%
\end{adjustbox}
\caption{Non-reasoning generation answer-quality results on the considered MHQA benchmarks.}
\label{tab:main_results_gpt41}
\end{table}

\begin{table}[t!] 
\centering
\begin{adjustbox}{width=0.98\columnwidth,center}
\begin{tabular}{@{}l c c c c c@{}}
\toprule
\textbf{Method} & \textbf{MuSiQue} & \textbf{2Wiki} & \textbf{MoreHopQA} & \textbf{FRAMES} & \textbf{Average} \\
\midrule
Direct Inference & 0.428 & 0.794 & 0.708 & 0.739 & 0.667 \\
CoT & 0.440 & 0.804 & 0.688 & 0.745 & 0.669 \\
\midrule
ReAct & 0.575 & 0.864 & 0.832 & 0.833 & 0.776 \\
IRCoT & 0.590 & 0.830 & 0.742 & 0.754 & 0.729 \\
\midrule
Coverage Anchor & 0.627 & \textbf{0.912} & \underline{0.860} & \textbf{0.860} & \underline{0.815} \\
Iterative Chain & \underline{0.628} & 0.884 & 0.830 & 0.834 & 0.794 \\
PAR$^2$-RAG & \textbf{0.639} & \underline{0.904} & \textbf{0.864} & \underline{0.858} & \textbf{0.816} \\
\bottomrule
\end{tabular}%
\end{adjustbox}
\caption{Reasoning generation answer-quality results on the considered MHQA benchmarks.}
\label{tab:main_results_gpto3}
\end{table}

Tables~\ref{tab:main_results_gpt41} and~\ref{tab:main_results_gpto3} show that PAR$^2$-RAG consistently delivers the strongest overall answer quality across benchmarks and generator settings. Under non-reasoning generation, PAR$^2$-RAG achieves the best performance on all four datasets. In particular, on MuSiQue, PAR$^2$-RAG yields its largest relative improvements: +42.4\% over ReAct and +23.5\% over IRCoT.

Under reasoning generation, PAR$^2$-RAG remains best on average. The largest relative gain over Coverage Anchor is +1.9\% on MuSiQue, while the largest relative gain over Iterative Chain is +4.1\% on MoreHopQA. The runner-up pattern also differs by regime: Iterative Chain is second-best under non-reasoning generation, while Coverage Anchor is second-best under reasoning generation. A plausible explanation is that the dominant bottleneck shifts with generator capability: with non-reasoning generation, stronger chain refinement helps more after retrieval, while with reasoning generation, broader evidence coverage becomes relatively more important.

\begin{table*}[t!] 
\centering
\begin{adjustbox}{width=0.98\textwidth,center}
\begin{tabular}{@{}l c c c c c c c c c c c c c c c@{}}
\toprule
\multirow{2}{*}{\textbf{Method}} & \multicolumn{3}{c}{\textbf{MuSiQue}} & \multicolumn{3}{c}{\textbf{2Wiki}} & \multicolumn{3}{c}{\textbf{MoreHopQA}} & \multicolumn{3}{c}{\textbf{FRAMES}} & \multicolumn{3}{c}{\textbf{Average}} \\
\cmidrule(lr){2-4} \cmidrule(lr){5-7} \cmidrule(lr){8-10} \cmidrule(lr){11-13} \cmidrule(lr){14-16}
& {NDCG} & {Recall} & {All Pass} & {NDCG} & {Recall} & {All Pass} & {NDCG} & {Recall} & {All Pass} & {NDCG} & {Recall} & {All Pass} & {NDCG} & {Recall} & {All Pass} \\
\midrule
ReAct & 0.596 & 0.581 & 0.240 & 0.724 & 0.683 & 0.386 & 0.896 & 0.891 & 0.782 & 0.717 & 0.671 & 0.355 & 0.733 & 0.707 & 0.441 \\
IRCoT & 0.611 & 0.692 & \underline{0.420} & 0.750 & \underline{0.795} & \underline{0.622} & 0.822 & 0.884 & 0.771 & \underline{0.804} & \underline{0.806} & \underline{0.565} & 0.747 & 0.794 & 0.595 \\
\midrule
Coverage Anchor & 0.610 & \underline{0.737} & \underline{0.428} & 0.708 & \underline{0.804} & 0.588 & 0.881 & \underline{0.944} & \underline{0.888} & 0.785 & 0.794 & 0.544 & 0.746 & \underline{0.820} & \underline{0.612} \\
Iterative Chain & \underline{0.615} & 0.691 & 0.392 & \underline{0.762} & 0.768 & 0.548 & \underline{0.898} & 0.895 & 0.790 & 0.785 & 0.794 & 0.545 & \underline{0.765} & 0.787 & 0.569 \\
PAR$^2$-RAG & \textbf{0.636} & \textbf{0.747} & \textbf{0.460} & \textbf{0.784} & \textbf{0.877} & \textbf{0.728} & \textbf{0.908} & \textbf{0.954} & \textbf{0.908} & \textbf{0.834} & \textbf{0.874} & \textbf{0.677} & \textbf{0.791} & \textbf{0.863} & \textbf{0.693} \\
\bottomrule
\end{tabular}%
\end{adjustbox}
\caption{Retrieval performance on the considered multi-hop QA benchmarks. We report NDCG, Recall, and All Pass for the retrieval stage. Best results are in bold and second-best are underlined.}
\label{tab:main_results_retrieval}
\end{table*}

Table~\ref{tab:main_results_retrieval} provides direct evidence for the mechanism. PAR$^2$-RAG achieves the best retrieval performance on Recall, NDCG, and All-Pass across all four datasets. These retrieval gains align with the answer-quality improvements. We hypothesize that ReAct and IRCoT are less competitive in this setting because iterative retrieval can suffer from \emph{early commitment}: reasoning starts before evidence coverage is sufficiently broad, and early low-recall steps can bias later retrieval. In contrast, PAR$^2$-RAG explicitly separates coverage expansion and chain refinement, which provides more controllable behavior under fixed retrieval budgets.

\subsection{Ablation Study}\label{sec:ablation}
\begin{figure*}[t!]
\centering
\includegraphics[width=\linewidth]{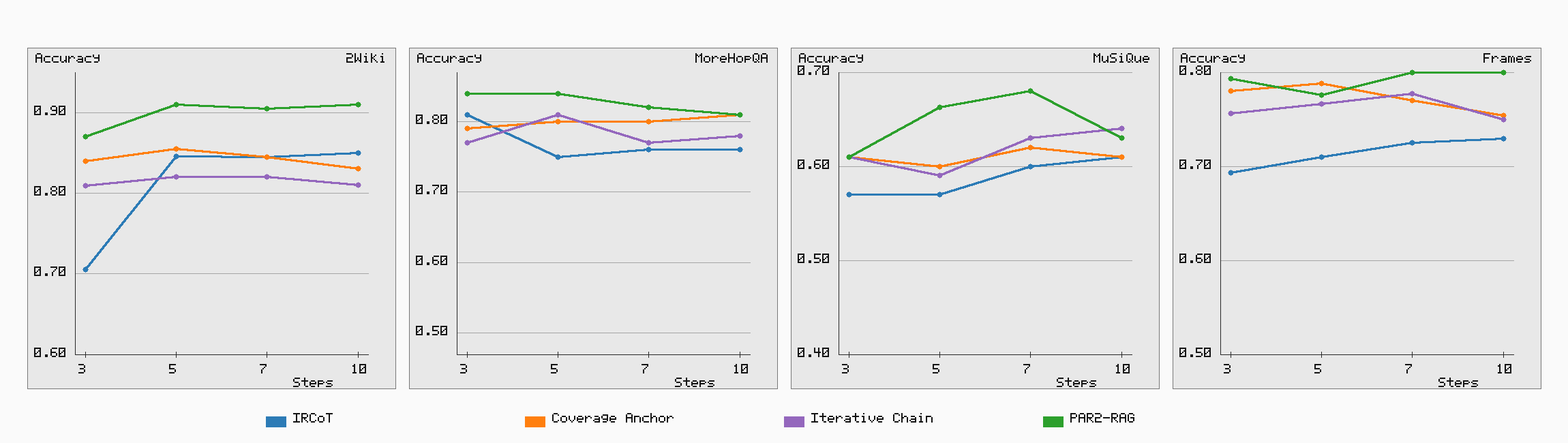}
\caption{Step robustness results on answer quality, each panel compares IRCoT, Coverage Anchor, Iterative Chain, and PAR$^2$-RAG across step counts $\\{3,5,7,10\\}$.}
\label{fig:step_robustness_lines}
\end{figure*}
We conduct ablations to analyze how robustness and where PAR$^2$-RAG's gains come from.

\paragraph{Ablation on GPT-5.2 Settings}
\begin{table}[t!]
\centering
\begin{adjustbox}{width=0.98\columnwidth,center}
\begin{tabular}{@{}l c c c c c@{}}
\toprule
\textbf{Method} & \textbf{MuSiQue} & \textbf{2Wiki} & \textbf{MoreHopQA} & \textbf{FRAMES} & \textbf{Average} \\
\midrule
\multicolumn{6}{c}{\textbf{OpenAI-GPT-5.2} Reasoning=None} \\
\midrule
IRCoT & 0.542 & 0.747 & 0.698 & 0.533 & 0.630 \\
Coverage Anchor & 0.603 & 0.785 & 0.757 & 0.707 & 0.713 \\
Iterative Chain & \underline{0.607} & \underline{0.827} & \underline{0.782} & \underline{0.721} & \underline{0.734} \\
PAR$^2$-RAG & \textbf{0.622} & \textbf{0.838} & \textbf{0.808} & \textbf{0.773} & \textbf{0.760} \\
\midrule
\multicolumn{6}{c}{\textbf{OpenAI-GPT-5.2} Reasoning=Medium} \\
\midrule
IRCoT & 0.622 & 0.807 & 0.682 & 0.590 & 0.675 \\
Coverage Anchor & \underline{0.661} & 0.859 & 0.845 & \underline{0.849} & 0.803 \\
Iterative Chain & {0.652} & \underline{0.879} & \textbf{0.864} & 0.829 & \underline{0.806} \\
PAR$^2$-RAG & \textbf{0.683} & \textbf{0.889} & \underline{0.862} & \textbf{0.857} & \textbf{0.823} \\
\bottomrule
\end{tabular}%
\end{adjustbox}
\caption{GPT-5.2 results across MHQA benchmarks with two reasoning configurations.}
\label{tab:main_results_gpt52_extension}
\end{table}

We evaluate PAR$^2$-RAG and considered RAG-based methods with GPT-5.2 model under two reasoning configurations (\texttt{none} and \texttt{medium}) for both reasoning and generation. The experimental results in Table~\ref{tab:main_results_gpt52_extension} show that PAR$^2$-RAG maintains clear gains over IRCoT and both module variants in both reasoning settings. Under reasoning=\texttt{none}, the largest relative gain over Coverage Anchor is +9.3\% on FRAMES, and the largest gain over Iterative Chain is also +7.2\% on FRAMES. Under reasoning=\texttt{medium}, the largest relative gain over Coverage Anchor is +3.5\% on 2Wiki, while the largest gain over Iterative Chain is +4.8\% on MuSiQue.

This extension provides a refined view of the runner-up hypothesis. At the average level, Iterative Chain remains the runner-up in both settings. However, when moving from reasoning=\texttt{none} to reasoning=\texttt{medium}, the gap between Iterative Chain and Coverage Anchor narrows substantially, and Coverage Anchor becomes competitive on multiple datasets. This trend is consistent with our hypothesis: as generation-time reasoning becomes stronger, broader evidence coverage contributes relatively more to final accuracy.

\paragraph{Ablation on Step Robustness}
Figure~\ref{fig:step_robustness_lines} shows step/sub-query scaling for all considered methods. The experimental results show the following observations: the performance of all methods generally improves from 3 to 5/7 steps/sub-queries, then saturates or declines at 10 steps on some datasets. We can tell that the proposed PAR$^2$-RAG remains the strongest overall performance, achieving the highest answer quality score at 12 of 16 benchmark-step points and the highest average performance across all benchmarks. 

The results suggests that PAR$^2$-RAG is robust in terms of maintaining high quality under budget changes, even when deeper search does not help uniformly. Variance-based stability also highlights method differences: IRCoT and Iterative Chain exhibit larger step-wise fluctuations than Coverage Anchor, indicating that retrieval-first anchoring yields more stable behavior under budget variation. Overall, these patterns support the adaptive retrieval control view: additional depth helps only when newly retrieved evidence remains on-path.

\section{Conclusions}
Finding high-quality evidence across multi-source corpora for multi-hop question answering remains challenging.
We proposed Planned Active Retrieval and Reasoning RAG (PAR$^2$-RAG), a two-stage framework that delays reasoning commitment until sufficient evidence coverage is achieved for MHQA. Across four MHQA benchmarks, PAR$^2$-RAG is consistently best for answer quality and achieves the strongest retrieval performance. Together with step-depth ablations, these results support a practical design rule for industry MHQA systems: use coverage-first anchoring before deep refinement, and prefer adaptive depth control to fixed large hop budgets.

\newpage
\appendix
\section{Appendix}\label{appendix:appendix}

\subsection{Dataset}\label{appendix:dataset}
\begin{table}[h!] 
\centering
\begin{adjustbox}{width=1.0\columnwidth,center}
\label{tab:dataset_stats}
\begin{tabular}{lrrr}
\toprule
\textbf{Dataset} & \textbf{Queries} & \textbf{Docs } & \textbf{Avg Text Length} \\
\midrule
MuSiQue          & 500                  & 19990                & 489.52                   \\
2Wiki            & 500                  & 10000                & 116.07                   \\
MoreHopQA         & 500                  & 9942                 & 132.44                   \\
FRAMES         & 500                  & 9942                 & 132.44                   \\
\midrule
\textbf{Average} & \textbf{500}         & \textbf{13311}       & \textbf{246.01}          \\
\bottomrule
\end{tabular}
\end{adjustbox}
\caption{Statistics of the multi-hop QA datasets. 'Query Number' and 'Docs Number' refer to the total count of queries and documents, respectively. 'Avg Text Length' is the average word count per document.}
\end{table}

\subsection{Evaluation Metrics}\label{appendix:metrics}

 \paragraph{Correctness Evaluation}
  To further evaluate the generated answer quality beyond string matching, we use a Correctness metric to measure the proportion of generated responses that are judged as
  correct by an LLM evaluator using a structured prompt template:
  \begin{equation}
  \mathrm{Correct} = \frac{\left|\{\,r \mid \text{LLM-Judge}(q, p, g) =
  \text{``yes''}\,\}\right|}{|P|}
  \end{equation}
  where \(q\)
  denotes the query, \(g\) denotes the ground-truth answer, and \(\text{LLM-Judge}(q, p, g)\)
  is a language model that evaluates whether the response \(p\) correctly answers query \(q\)
  given ground truth \(g\). The evaluation uses GPT-4 with a predefined prompt template in Prompt~\ref{fig:prompt_incoming_questions}, which outputs binary judgments, with responses containing
  "yes" (case-insensitive) scored as 1 and others as 0.
  \newtcolorbox{promptbox_correct}{
    colback=white,      
    coltext=black,      
    fonttitle=\bfseries,
    coltitle=white,
    boxrule=0pt,        
    arc=0mm,            
    boxsep=5pt,
    left=6pt,
    right=6pt,
    top=6pt,
    bottom=6pt,
    title={Prompt for Correctness judgment} 
}
\begin{figure*}[ht]
\centering
\begin{adjustbox}{width=0.9\textwidth, center}
\begin{promptbox_correct}
You are a helpful research assistant. Your task is to evaluate an LLM's answer against a ground-truth answer and decide whether the ground-truth content is present in the model's response.

\vspace{0.75em}
\textbf{Instructions:}
\begin{enumerate}[label=\arabic*., leftmargin=*, itemsep=2pt, topsep=2pt]
    \item Carefully compare the \emph{Predicted Answer} with the \emph{Ground-Truth Answer}.
    \item Judge based on substance and equivalence of meaning; do not require identical wording unless wording is crucial to meaning.
    \item Make a binary decision on whether the vital facts of the ground-truth are contained in the predicted answer.
\end{enumerate}

\vspace{0.75em}
\textbf{Input Data:}
\begin{promptlisting}
Question: {query}
Predicted Answer: {response}
Ground-Truth Answer: {answer}
\end{promptlisting}

\vspace{0.75em}
\textbf{Output Format:}

Provide your final evaluation in the following format:
\begin{promptlisting}
Explanation: <brief rationale for the decision>
Decision: <yes|no>
\end{promptlisting}

\vspace{0.75em}
\noindent\textbf{Output:}

\end{promptbox_correct}
\end{adjustbox}
\caption{Prompt for correctness judgment.}
\label{fig:prompt_incoming_questions}
\end{figure*}

  \paragraph{NDCG}
  The Normalized Discounted Cumulative Gain (NDCG) metric
  evaluates the ranking quality of retrieved documents by
  considering both relevance and position:
  \begin{equation}
  \mathrm{NDCG@k} = \frac{\mathrm{DCG@k}}{\mathrm{IDCG@k}}
  \end{equation}
  where DCG@k is the Discounted Cumulative Gain at rank k,
  calculated as:
  \begin{equation}
  \mathrm{DCG@k} = \sum_{i=1}^{k} \frac{2^{\text{rel}_i} -
  1}{\log_2(i + 1)}
  \end{equation}
  and IDCG@k is the ideal DCG@k obtained by sorting all relevant
   documents in descending order of relevance. Here,
  $\text{rel}_i$ represents the relevance score of the document
  at position $i$, with binary relevance where documents
  containing ground-truth context receive a score of 1 and others receive 0.

 \paragraph{Retrieval Recall}
The Retrieval Recall metric measures the proportion of queries
for which at least one ground-truth context is successfully
retrieved in the top-k results:
  \begin{equation}
  \mathrm{Recall@k} = \frac{\left|q \mid \text{GT}_q \cap
  \text{Retrieved@k}_q \neq \emptyset\right|}{|Q|}
  \end{equation}
  where $Q$ denotes the set of all queries, $\text{GT}_q$
  represents the set of ground-truth for query $q$, and
  $\text{Retrieved@k}_q$ represents the set of documents in the top-k retrieved results for query $q$. Each query receives a binary score of 1 if any ground-truth appears in the retrieval results, and 0 otherwise.

\subsection{Prompts}\label{appendix:prompt}
\newtcolorbox{promptbox_planner}{
    colback=white,      
    coltext=black,      
    fonttitle=\bfseries,
    coltitle=white,
    boxrule=0pt,        
    arc=0mm,            
    boxsep=5pt,
    left=6pt,
    right=6pt,
    top=6pt,
    bottom=6pt,
    title={Prompt for Planner} 
}
\begin{figure*}[ht]
\centering
\begin{adjustbox}{width=0.9\textwidth, center}
\begin{promptbox_planner}
You are a helpful research assistant. Given a query, come up with a set of database searches to perform to best answer the query. 
Output 5 terms to query for.

Format your response as JSON (No code snippet) with a list of database searches needed to answer the query. Each search should include:
\vspace{0.75em}
\begin{enumerate}[label=\arabic*., itemsep=2pt, topsep=4pt, leftmargin=*, font=\color{black}]
    \item \textbf{Reason:} A brief explanation of why this search is necessary.
    \item \textbf{Query:} The exact search term to use.
\end{enumerate}
\vspace{0.75em}
\textbf{Example output:}
\begin{promptlisting}
{
  "searches": [
    {
      "reason": "Identify the best Caribbean destinations for surfing in April.",
      "query": "best Caribbean surfing spots April"
    },
    {
      "reason": "Find hiking trails in the Caribbean suitable for April vacations.",
      "query": "hiking trails Caribbean April"
    }
  ]
}
\end{promptlisting}

\end{promptbox_planner}
\end{adjustbox}
\caption{Prompt for planner.}
\label{fig:prompt_planner}
\end{figure*}

\newtcolorbox{promptbox_searcher}{
    colback=white,      
    coltext=black,      
    fonttitle=\bfseries,
    coltitle=white,
    boxrule=0pt,        
    arc=0mm,            
    boxsep=5pt,
    left=6pt,
    right=6pt,
    top=6pt,
    bottom=6pt,
    title={Prompt for Searcher} 
}
\begin{figure*}[ht]
\centering
\begin{adjustbox}{width=0.9\textwidth, center}
\begin{promptbox_searcher}
You are a database search interface. Your ONLY responsibility is to return the FIVE database results exactly as provided, without any modification.

\vspace{0.75em}
\textbf{Strict Rules:}
\vspace{0.75em}
\begin{enumerate}[label=\arabic*., itemsep=2pt, topsep=4pt, leftmargin=*, font=\color{black}]
    \item Output must consist ONLY of the raw database results. 
    \item Preserve the original wording, spelling, punctuation, capitalization, line breaks, and formatting of the database entries. 
    \item Do NOT summarize, rephrase, explain, interpret, or add commentary. 
    \item Do NOT add introductions, conclusions, or transitional text. 
    \item Do NOT merge, reorder, or alter results beyond their original sequence. 
    \item If multiple entries are returned, output them in exactly the same order and format as they are given by the database. 
    \item MUST output all five retrieved entries.
\end{enumerate}

\vspace{0.75em}
\textbf{Example (for illustration only --- do not generate similar text unless the database returns it):}
\begin{promptlisting}
[Document1.txt]
Document1 is an example entry from the database.

[Document2.txt]
Document2 is other entry in the database.

[Document3.txt]
Document3 is other entry in the database.

[Document4.txt]
Document4 is other entry in the database.

[Document5.txt]
Document5 is other entry in the database.
\end{promptlisting}

\end{promptbox_searcher}
\end{adjustbox}
\caption{Prompt for searcher.}
\label{fig:prompt_searcher}
\end{figure*}

\newtcolorbox{promptbox_writer}{
    colback=white,
    coltext=black,
    fonttitle=\bfseries,
    coltitle=white,
    boxrule=0pt,
    arc=0mm,
    boxsep=5pt,
    left=6pt,
    right=6pt,
    top=6pt,
    bottom=6pt,
    title={Prompt for Writer}
}

\begin{figure*}[ht]
\centering
\begin{adjustbox}{width=0.9\textwidth, center}
\begin{promptbox_writer}
You are a senior researcher tasked with providing a comprehensive answer to a research query.
You will be provided with the original query, and initial research done by a research assistant. 

\textbf{DO NOT WRITE A SUMMARY}, \textbf{directly provide a complete, accurate answer} to the original question.

\vspace{0.75em}
\textbf{Output Format}

Format your response as \textbf{JSON (No code snippet)} with:
\begin{itemize}[leftmargin=*, itemsep=2pt, topsep=2pt]
    \item \texttt{answer}: An answer that directly addresses the research query using all available evidence.
\end{itemize}

\textbf{Example output:}
\begin{promptlisting}
{
  "answer": "Based on the research findings, the Caribbean in April offers exceptional conditions for outdoor activities..."
}
\end{promptlisting}

\end{promptbox_writer}
\end{adjustbox}
\caption{Prompt for Writer.}
\label{fig:prompt_writer}
\end{figure*}

\subsection{Implementation Details}\label{appendix:implementation}
All main experiments use OpenSearch TopK=100 and rerank TopK=5. Unless otherwise stated, the step/sub-query budget is 5 and retrieval summary is disabled (\texttt{self.nosummary\_3}). The retriever and reranker use \texttt{openai.o4-mini-2025-04-16}. Main settings are:
\begin{itemize}
    \item Non-reasoning generation setting: reasoning model \texttt{openai.gpt-4.1-2025-04-14}, generation model \texttt{openai.gpt-4.1-2025-04-14}.
    \item Reasoning generation setting: reasoning model \texttt{openai.gpt-4.1-2025-04-14}, generation model \texttt{openai.o3-2025-04-16}.
    \item Extension setting: reasoning and generation models \texttt{openai.gpt-5.2-2025-12-11} with reasoning mode variations.
\end{itemize}

\subsection{Ablation Details}\label{appendix:ablation}
\paragraph{Step/sub-query without reasoning generation (300-sample study)}
For PAR$^2$-RAG, correctness on MuSiQue increases from 0.610 (3 steps) to 0.680 (7 steps), then drops to 0.630 (10 steps). FRAMES increases from 0.793 (3) to 0.800 (7--10). Similar non-monotonic trends appear for IRCoT, Multi-step RAG, and Query Decomposition.

\paragraph{Step/sub-query with reasoning generation (300-sample study)}
With reasoning generation, PAR$^2$-RAG reaches 0.940 on 2Wiki at 7 steps and 0.700 on MuSiQue at 10 steps, while FRAMES peaks at 0.866 at 7 steps. The overall trend remains non-monotonic, reinforcing the need for adaptive stopping.

\subsection{Additional Results}\label{appendix:results}
\paragraph{GPT-5.2 extension}
With reasoning mode set to none, PAR$^2$-RAG reports 0.838 (2Wiki), 0.808 (MoreHopQA), 0.622 (MuSiQue), and 0.773 (FRAMES). With reasoning mode set to medium, PAR$^2$-RAG reports 0.889, 0.862, 0.683, and 0.857, respectively.

\end{document}